\documentclass[11pt]{article}

\usepackage[final]{acl}

\usepackage{times}
\usepackage{latexsym}

\usepackage[T1]{fontenc}

\usepackage[utf8]{inputenc}

\usepackage{inconsolata}

\usepackage{graphicx}
\usepackage{hyperref}
\usepackage{url}
\usepackage{siunitx}
\usepackage{booktabs}       
\usepackage{amsfonts}       
\usepackage{xcolor}
\usepackage{nicefrac}       
\usepackage[activate=true,
final,
tracking=false,   
kerning=true,
spacing=true,
factor=1050,
stretch=30,
shrink=30]{microtype}      

\usepackage{enumitem}
\usepackage{pifont}   

\usepackage{xspace}

\usepackage{url}
\usepackage{mathtools}
\usepackage{amsmath}
\usepackage{amsthm}
\usepackage{dsfont}
\usepackage{amssymb}
\usepackage{wrapfig}
\usepackage{subcaption}
\usepackage{bbm}
\usepackage{bm}
\usepackage{multirow}
\usepackage{booktabs}
\usepackage{xspace}

\usepackage[capitalize]{cleveref}
\crefname{page}{page}{pages}
\crefname{footnote}{footnote}{footnotes}   
\crefname{equation}{equation}{equations}   
\crefname{corollary}{Corollary}{Corollaries}  
\crefname{line}{line}{lines}               
\crefrangeformat{line}{lines #3#1#4--#5#2#6}
\crefname{lstlsting}{Listing}{Listings}   
\crefname{section}{\S}{\S\S}
\Crefname{section}{\S}{\S\S}    
\crefformat{section}{#2\S#1#3}  
\Crefformat{section}{#2\S#1#3}
\crefrangeformat{section}{\S\S#3#1#4--#5#2#6}
\Crefrangeformat{section}{\S\S#3#1#4--#5#2#6}
\crefmultiformat{section}{\S#2#1#3}{ and~\S#2#1#3}{, \S#2#1#3}{ and~\S#2#1#3}
\Crefmultiformat{section}{\S#2#1#3}{ and~\S#2#1#3}{, \S#2#1#3}{ and~\S#2#1#3}
\crefrangemultiformat{section}{\S\S#3#1#4--#5#2#6}{ and~\S\S#3#1#4--#5#2#6}{, \S\S#3#1#4--#5#2#6}{ and~\S\S#3#1#4--#5#2#6}
\Crefrangemultiformat{section}{\S\S#3#1#4--#5#2#6}{ and~\S\S#3#1#4--#5#2#6}{, \S\S#3#1#4--#5#2#6}{ and~\S\S#3#1#4--#5#2#6}
\usepackage{tcolorbox}
\usepackage{lipsum} 
\usepackage{tabularx}
\usepackage{algorithm}  
\usepackage[endLComment=]{algpseudocodex}

\definecolor{myDeepYellow}{rgb}{0.9412, 0.6902, 0.302}
\definecolor{myYellow}{rgb}{0.9765, 0.8824, 0.7255}
\definecolor{myBlue}{rgb}{0.6353, 0.7686, 0.8627}
\definecolor{darkYellow}{rgb}{0.8, 0.6, 0.1} 
\definecolor{lightYellow}{rgb}{0.98, 0.92, 0.6} 

\def\docommandbetter#1 {\colorbox{myBlue!80}{#1} \let\next\argii}
\def\argii{\let\next\docommandbetter}

\newcommand{\styleToken}[1]{\textcolor{blue}{#1}}
\newcommand{\pitchToken}[1]{\textcolor{orange}{#1}}

\makeatletter
\newcommand{\smartperiod}{\@ifnextchar.{}{.\@\xspace}}
\newcommand{\smartcomma}{\@ifnextchar.{}{,}\xspace}
\makeatother
\newcommand{\latin}[1]{#1}  

\newcommand{\ie}{\latin{i.e.}\smartcomma}

\newcommand{\xmark}{\ding{55}}
\newcommand{\cmark}{\ding{51}}

\usepackage{tabularx}
\renewcommand{\arraystretch}{1.2}              
\newcolumntype{Y}{>{\raggedright\arraybackslash}X} 

\definecolor{asuffix}{RGB}{178,39,238}
\definecolor{redbrown}{RGB}{165,42,42}

\definecolor{grayblack}{rgb}{0.3, 0.3, 0.3} 

\title{Seeing is Believing: Emotion-Aware Audio-Visual Language Modeling for Expressive Speech Generation}

\author{
  Weiting Tan$^{\spadesuit}$\thanks{\ \ Work was done during an internship at Meta AI.} \quad
  Jiachen Lian$^{\heartsuit}$ \quad
  Hirofumi Inaguma$^{\heartsuit}$\\
  \textbf{Paden Tomasello}$^{\heartsuit}$ \quad
  \textbf{Philipp Koehn}$^{\spadesuit}$ \quad
  \textbf{Xutai Ma}$^{\heartsuit}$ \\
  $^{\spadesuit}$Johns Hopkins University \quad
  $^{\heartsuit}$Meta AI Research
}

\begin{document}
\maketitle
\begin{abstract}
We present an Audio-Visual Language Model (AVLM) for expressive speech generation by integrating full-face visual cues into a pre-trained expressive speech model. We explore multiple visual encoders and multimodal fusion strategies during pre-training to identify the most effective integration approach. Subsequent fine-tuning on emotion recognition and expressive dialogue tasks yields substantial gains over speech-only baselines (e.g., $+5$ F1 in emotion recognition). AVLM highlights the value of expressive visual information in guiding speech generation and offers a foundation for end-to-end multimodal conversational systems.\footnote{All experiments, data collection, and processing activities were conducted by JHU. Meta was involved solely in an advisory role and no experiments, data collection or processing activities were conducted on Meta infrastructure. Code accessible at:\url{https://github.com/steventan0110/AVLM}}
\end{abstract}

\section{Introduction}

Expressive speech generation are supported by high-quality speech tokenization/codec \cite{encodec, jenrungrot2023lmcodeclowbitratespeech, du2024cosyvoice2scalablestreaming, nguyen2024spiritlm}, emotionally-nuanced speech datasets \cite{chu2024qwen2audiotechnicalreport, stepaudio, kimiaudio}, and effective emotion-aware representation learning \cite{wang2023learningemotionalrepresentationsimbalanced,tang2024edttsmultiscaleemotionmodeling}. While recent advances in speech modeling have greatly improved the quality and controllability of synthesized speech, most methods rely solely on audio input and overlook the rich expressive cues available in the visual modality. 

Human communication, however, is inherently multimodal. Visual signals—such as facial expressions, head gestures, and eye movements—are tightly intertwined with vocal expression, offering critical paralinguistic information that reflects emotion and intent. This raises a central question: \textit{can we enhance expressive speech generation by incorporating these visual cues, enabling the model to produce responses that are not only fluent but also emotionally resonant and contextually adaptive?}

To address this question, we propose developing an emotion-aware audio-visual perceptual model for expressive speech generation. Although prior work has explored lip reading, audio-visual speech recognition (AVSR), and emotion recognition \cite{avhubert, Hong2023WatchOL, han-etal-2024-xlavs, yeo2025zeroavsrzeroshotaudiovisualspeech, yeo2025mmsllamaefficientllmbasedaudiovisual, unified_speech_recognition, Busso2008IEMOCAPIE}, these efforts are often task-specific. AVSR methods, for example, improve transcription in noisy settings by focusing on the lip region, but they generally ignore non-semantic yet affective visual signals. Similarly, emotion recognition benchmarks like IEMOCAP \cite{Busso2008IEMOCAPIE} are designed for classification rather than generation, limiting their utility in expressive synthesis tasks. Consequently, current systems fall short in capturing the full emotional dynamics present in audio-visual communication.


\begin{figure}[t]
  \centering
  \includegraphics[width=0.9\linewidth]{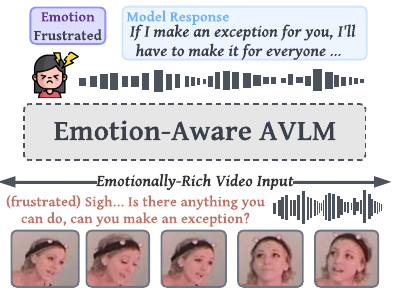}
  \caption{Our Audio-Visual Language Model (AVLM) perceives an audio-visual input, determine the emotion of the speaker and generate an expressive response.}
  \label{figure::teaser}
  \vspace{-1em}
\end{figure}

In this work, we propose an Audio-Visual Language Model (AVLM) that integrates full-face visual cues into a pre-trained expressive Speech Language Model (SpeechLM) to support emotionally rich speech generation. Our training follows a two-stage framework. First, we introduce a modality fusion module to align visual representations with SpeechLM’s latent space, systematically evaluating visual encoders and fusion strategies during pre-training. Second, we fine-tune the AVLM for emotion recognition and dialogue generation using synthetic data derived from IEMOCAP. To construct expressive conversations, we rewrite dialogue responses using GPT-4 \cite{openai2024gpt4} and synthesize audio with Step-Audio \cite{stepaudio}, an expressive TTS system supporting voice cloning and natural language instruction.

Our approach effectively incorporates a wider range of visual information beyond lip movements.  It achieves lower perplexity during pre-training and improves AVSR performance by reducing Word Error Rate (WER) over 1 point under clean and noisy conditions. When fine-tuned for expressive speech generation and emotion recognition, our visually enhanced AVLM consistently outperforms the speech-only counterpart (by more than 5 F1-score), demonstrating higher classification accuracy and generating more expressive outputs. In summary, we strengthen the connection between visual and audio modality for expressive generation and show that our novel AVLM could be a strong foundation for building emotionally intelligent end-to-end conversational agents.

\section{Related Work} 

\subsection{Audio-Visual Speech Recognition} 
AVSR leverages both audio and visual modalities for robust speech transcription. Datasets like LRS3 \cite{afouras2018lrs3tedlargescaledatasetvisual} and VoxCeleb2 \cite{voxceleb_two} have driven recent progress. AV-HuBERT \cite{avhubert} aligns modalities via self-supervised learning, extended to multilingual settings by XLAVS-R \cite{han-etal-2024-xlavs}. Unified-Speech \cite{unified_speech_recognition} incorporates auxiliary tasks to improve multi-modal alignment, while LLaMa-AVSR \cite{cappellazzo2025largelanguagemodelsstrong} and MMS-LLaMa \cite{yeo2025mmsllamaefficientllmbasedaudiovisual} employ LLMs for improved transcription quality.

\subsection{Audio-Visual Emotion Recognition}
Multimodal emotion recognition has been supported by datasets such as IEMOCAP \cite{Busso2008IEMOCAPIE}, RECOLA \cite{recola}, CREMA-D \cite{cao2014crema}, MSP-IMPROV \cite{busso2017msp}, RAVDESS \cite{livingstone2018ravdess}, and Aff-Wild \cite{affwild}. Labels range from discrete classes to continuous dimensions like Valence, Arousal, and Dominance \cite{mehrabian1974approach}. Based on these datasets, fusion methods \cite{hsemotion, praveen2023er, ma2024transformer} have been proposed to improve multimodal emotion recognition, though they are limited to classification or regression settings.

\subsection{Speech Language Models}
Speech Language Models (SpeechLMs) are language models trained on large text and speech datasets, typically pre-trained on text and fine-tuned with speech-text or speech-only data \citep{asr_llama, yu2023connectingspeechencoderlarge, tan2024ssralignmentawaremodalityconnector, speechgpt, tang2024salmonn, chu2024qwen2audiotechnicalreport}. 


More recently, directly encoding speech as tokens and integrating them into pre-trained LLMs has gained traction for its scalability and ease of expressive speech synthesis \cite{nguyen2024spiritlm, stepaudio, minmo, kimiaudio}. SpiritLM \cite{nguyen2024spiritlm}, for instance, enhances expressiveness by incorporating style and pitch tokens \cite{kharitonov-etal-2022-text, sonar_expressive}, while Step-Audio \cite{stepaudio} leverages a linguistic tokenizer trained on Paraformer outputs \cite{paraformer}. The focus on expressivity in these models makes them strong candidates for our visual integration.

\section{Motivate the Integration of Visual Cues: Audio-Visual Emotion Recognition}\label{sec::iemocap_recognition}
Before diving into Expressive Audio-Visual Language Modeling, we first motivate the study by showing why visual cues, besides its semantic correlation to speech in lip-reading, could be useful. We choose the emotion recognition benchmark, IEMOCAP \cite{Busso2008IEMOCAPIE}, and compare the performance of existing audio-only baselines and our simple audio-visual classification model.

IEMOCAP is made up of videos of two speakers' conversation with strong emotions. We follow prior benchmark EmoBox \cite{Ma2024EmoBoxMM} to train and evaluate models with their released data splits\footnote{~\url{https://github.com/emo-box/EmoBox}}. We follow the findings of EmoBox to use Whisper \cite{whisper} as the audio encoder. For visual information, we directly encode frames with pre-trained Open-MAGVIT2 \cite{magvit2,luo2025openmagvit2opensourceprojectdemocratizing} encoder\footnote{~\url{https://github.com/TencentARC/SEED-Voken}}, which has achieved impressive image and video reconstruction performance.

\begin{table}[t]
  \small
  \centering
  \resizebox{0.98\linewidth}{!}{
  \begin{tabular}{l c c c c}
  \toprule
  \textbf{Model (Modality)} & \textbf{UA (\%)} $\uparrow$ & \textbf{WA (\%)} $\uparrow$& \textbf{F1 (\%)}$\uparrow$ \\
  \midrule
  EmoBox (Speech)          & 73.5  & 72.9   & 73.1 \\
  SenseVoice-L (Speech)    & 73.9  & 75.3   & 73.2 \\
  \midrule
  Ours (Visual)         & 77.9  & 79.1   & 78.7 \\
  Ours (Speech+Visual) & \textbf{88.2}  & \textbf{89.2}   & \textbf{88.5} \\
  \bottomrule
  \end{tabular}}
  \vspace{-0.5em}
  \caption{Comparison of emotion recognition models leveraging different modalities. The performance is evaluated over unweighted average accuracy (UA), weighted average accuracy (WA), and macro F1 score (F1) following SenseVoice \cite{sensevoice}.}
  \label{table::audio_visual_emo_classification}
  \vspace{-1em}
\end{table}

To train our classifier, we encode visual and speech input into features, adapt them through lightweight encoders, and feed them into a Transformer-Encoder model \cite{transformer_neurips}. The features are then pooled and passed into a Feedforward network to predict the emotion label. For details of our model setup, please refer to \cref{app::iemocap_emo_recognition}. As shown in \cref{table::audio_visual_emo_classification}, we compare our classification model with state-of-the-art speech-only models, EmoBox \cite{Ma2024EmoBoxMM} and SenseVoice \cite{sensevoice}. We find that, by incorporating visual modality, our model achieves superior performance. 


We perform an ablation study by grouping samples based on the prediction outcomes of each modality. As shown in \cref{table::emo_classification_ablation}, the fusion model makes more correct predictions via an \textit{ensembling mechanism}. Notably, only a small number of samples are correctly predicted solely by the fusion model (119) or by a single-modality model alone (38). In most cases, the fusion model aligns with the correct single-modality prediction—646 times with speech-only and 537 times with visual-only model—indicating effective modality selection.

These findings indicate that the visual modality provides complementary non-semantic information. Effectively leveraging such visual signals holds promise for enhancing the expressiveness and emotional alignment of speech generation systems.

\begin{figure*}[t]
  \centering
  \includegraphics[width=0.95\linewidth]{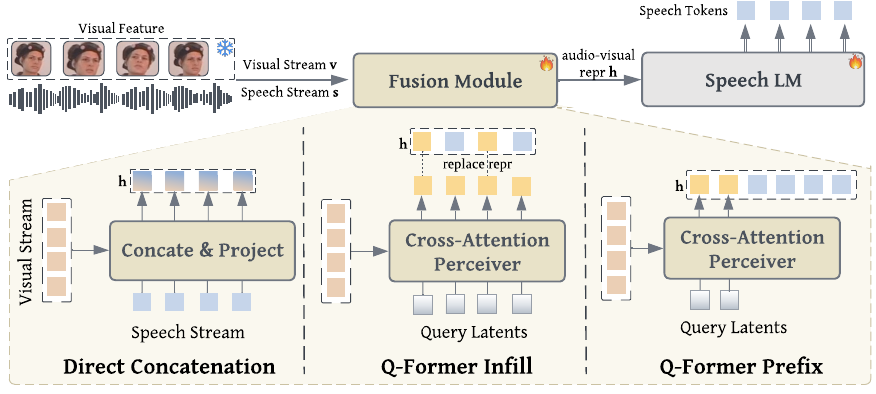}
  \vspace{-0.5em}
  \caption{Illustration of the modality fusion strategies explored in the AVLM \textbf{pre-training} stage. The fusion module is used to align audio and visual features and feed joint representations to SpeechLM for speech token prediction.}
  \label{figure::modality_fusion}
\end{figure*}

\section{Audio-Visual Language Modeling}\label{sec::avlm}
To support expressive speech generation, we propose Audio-Visual Language Modeling and explore modality fusion strategies in \cref{sec::avlm_pretrain}. We then discuss our fine-tuning strategy for emotion recognition and expressive speech generation in \cref{sec::emo_gen}.

\subsection{AVLM Pre-training}\label{sec::avlm_pretrain}
To integrate visual modality into a pre-trained SpeechLM, we leverage self-supervised learning on raw video data (\textit{to distinguish this stage from future fine-tuning experiments, we call this phase AVLM pre-training}). Since the base SpeechLM is already trained on speech and text tokens and employs a Transformer Decoder-only architecture, we aim to inject and align visual information into the existing speech latent space while maintaining next-token prediction as the training objective.

Given speech features $\bm s = (s_1, s_2, \cdots, s_T)$ and visual features $\bm v = (v_1, v_2, \cdots, v_T)$, we propose three audio-visual fusion strategies to obtain audio-visual representations $\bm h$, as visualized in \cref{figure::modality_fusion}.

\textsc{Direct Concat} fuses modalities by simply concatenating the time-aligned audio and visual features, which are then projected through a two-layer feedforward network (FFN). The resulting feature is subsequently fed into SpeechLM to generate the speech token sequence $\bm{s}$.

Both \textsc{Q-Former Infill} and \textsc{Q-Former Prefix} leverage the Q-Former module~\cite{blip2}, which uses a set of query latents to retrieve relevant information from the visual stream. These query latents are initialized as learnable parameters and refined via a cross-attention mechanism that attends to the visual features.

\begin{table}[t]
  \centering
  \small
  \begin{tabular}{c|c|c|c}
  \toprule
  Fusion & Speech & Visual & \# Samples \\
  \midrule
  \xmark & \cmark & \cmark & 38 \\
  \cmark & \xmark & \xmark & 119 \\
  \cmark & \cmark & \xmark & \textbf{646} \\
  \cmark & \xmark & \cmark & \textbf{537} \\
  \bottomrule
  \end{tabular}
  \vspace{-0.5em}
  \caption{Analysis of classification correctness across different models. \cmark\ indicates correct prediction while \xmark\ indicates incorrect prediction.}
  \label{table::emo_classification_ablation}
  \vspace{-1em}
\end{table}

Formally, given a visual stream $\bm{v}$ and a sequence of query latents $\bm{q} = (q_1, q_2, \cdots, q_{|\bm{q}|})$, we first apply sinusoidal positional embeddings to the query latents. They are then passed through a cross-attention module, where each layer contains learnable projection matrices $\bm{W}_\text{Q}, \bm{W}_\text{K}, \bm{W}_\text{V} \in \mathbb{R}^{d \times d}$. Here, $d$ is the shared dimensionality of the query, visual, and speech representation spaces. The query representation is computed using the standard cross-attention formula:
\begin{equation}
  \label{eq::cross_attn}
  \bm{z} = \text{softmax}\left(\frac{(\bm{q} \bm{W}_\text{Q}) (\bm{v} \bm{W}_\text{K})^T}{\sqrt{d}}\right) \cdot (\bm{v} \bm{W}_\text{V})
\end{equation}

For \textsc{Q-Former Infill}, we set the number of query latents to match the length of the speech sequence, i.e., $|\bm{z}| = |\bm{q}| = |\bm{s}|$, and randomly replace a portion of the speech representations with query latents. This approach is motivated by AV-HuBERT~\cite{avhubert}, where modality dropout encourages the model to learn robust audio-visual semantic alignment. In \textsc{Q-Former Prefix}, we \textbf{dynamically determine the number of query latents} based on the number of visual frames and use them as a prefix prepended to the speech representations.\footnote{In practice, we downsample the visual frames by a factor of $5\times$.} To ensure the model attends to the visual input, we additionally apply attention masking to some speech positions (see \cref{sec::exp_avlm_pretrain} for details).

To summarize, the audio-visual representation $\bm h$ is obtained with the following equations:
\begin{equation}
  \left\{
  \begin{aligned}
    \bm{h}_{i} &= \mathrm{FFN} \left([\bm{s}_i \circ \bm{v}_i]\right) 
    && \hspace{-1.6em} \textsc{(CONCAT)} \\
    \bm{h}_i &= \bm r_i \cdot \bm{z}_i + \left(1 - \bm r_i \right) \cdot \bm{s}_i 
    && \hspace{-0.2em} \textsc{(INFILL)} \\
    \bm{h}_i &= \bm{z}_i \ \text{if } i < |\bm{z}| \ \text{else } \bm{s}_i
    && \hspace{-0.5em} \textsc{(PREFIX)}
  \end{aligned}
  \right.
  \label{eq::fusion_eq}
\end{equation}
where $\bm r_i$ is 1 at position $i$ where we replace the speech with a visual representation and 0 otherwise. $\circ$ denotes concatenation along feature dimension. 

Subsequently, our AVLM is pre-trained with the negative log-likelihood (NLL) loss where parameters of the SpeechLM ($\theta_{\mathrm{lm}}$) and fusion module ($\phi_{\mathrm{fusion}}$) are both updated.
\begin{equation}
\mathcal{L}_{\mathrm{pretrain}} = - \sum_{t=1}^{T} \log P(s_t \mid \mathbf{h}_{<t}; \theta_{\mathrm{lm}}, \phi_{\mathrm{fusion}})
\end{equation}
In our experiments, we find that \textsc{Q-Former Prefix} achieves the best performance. Therefore, we adopt the \textsc{Q-Former Prefix} AVLM as the base model for fine-tuning in the next section.

\begin{figure*}[t]
  \centering
  \includegraphics[width=0.9\linewidth]{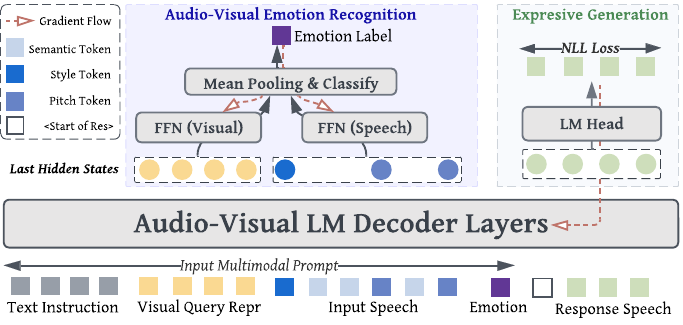}
  \caption{AVLM Multi-task \textbf{Fine-tuning} Objectives. The AVLM receives a multimodal input prompt and generates a speech response. An auxiliary emotion recognition module is trained based on transformed visual \& speech features.}
  \vspace{-1em}
  \label{figure::avlm_emo_gen}
\end{figure*}

\subsection{AVLM Fine-tuning for Emotion Recognition and Expressive Generation}\label{sec::emo_gen}
Given the pre-trained AVLM from \cref{sec::exp_avlm_pretrain}, we fine-tune it on expressive audio-visual conversations annotated with an emotion label. The dataset is derived from IEMOCAP, providing us with an input audio-visual pair, an emotion label, and the output speech. The construction details are shown in the experiment sections and \cref{app::dataset_details}, and some example data entries are provided in \cref{table::all-processed-example}.

\begin{tcolorbox}[
  colback=brown!5!white,
  colframe=brown!75!black,
  title=Instruction-Tuning Prompt,
  fonttitle=\footnotesize\bfseries,
  sharp corners,
  boxrule=0.8pt,
  left=2mm, right=2mm, top=1mm, bottom=1mm,
]
\footnotesize
\ttfamily
\noindent
<Text Instruction> ($\bm x_1, \bm x_2, \cdots$): {\color{gray}Perceive the given visual and audio input. Determine the emotion of the speaker and continue the dialogue with the same emotion}.\\[2pt]
<Visual Input>: {\color{yellow!80!black} $\bm{z}_1, \bm{z}_2, \cdots, \bm{z}_N$}\\
<Input Audio>: {\color{blue!80!black} $\bm{s}_1, \bm s_2, \cdots, \bm s_T$}\\
<Emotion>: {\color{asuffix!80!black}$\bm e$ (Happy, Angry, ...)}\\
<Speech Response>: {\color{green!50!black}$\bm y_1, \bm y_2, \cdots, \bm y_M$}
\end{tcolorbox}

As shown in \cref{figure::avlm_emo_gen} and colorbox above, we use a multimodal prompt for expressive speech generation. Specifically, we extract visual query representations $\bm z = \{\bm z_1, \cdots, \bm z_N\}$ from the input visual features and use them as a prefix to the input speech tokens $\bm s = \{\bm s_1, \cdots, \bm s_T\}$, with $N < T$ due to the compression from \textsc{Q-Former}. We then insert the emotion label $\bm e$ to guide generation, followed by the target response speech tokens $\bm y = \{\bm y_1, \cdots, \bm y_M\}$. 

To fine-tune the model, we minimize the negative log-likelihood (NLL) of the target response tokens $\bm{y} = \{y_1, \ldots, y_M\}$, conditioned on the multimodal prompt comprising the instruction $\bm{x}$, visual prefix $\bm{z}$, input speech $\bm{s}$, and emotion label $\bm{e}$:
\begin{equation}
\mathcal{L}_{\text{finetune}} = - \sum_{t=1}^{M} \log P\left( y_t \mid \left[ \bm{x} \mathbin\Vert \bm{z} \mathbin\Vert \bm{s} \mathbin\Vert \bm{e} \right] ; \theta_{\mathrm{lm}}\right)
\end{equation}
Here, $\mathbin\Vert$ denotes the concatenation of input components along the sequence dimension.

\paragraph{Auxiliary Emotion Recognition Task}
For expressive conversation synthesis using IEMOCAP, we ensure that both input and response speech share the same emotional tone. This is possible as IEMOCAP uses scripted dialogue where speakers of each conversation share similar emotions.

Since autoregressive emotion prediction is unreliable—due to limited data and hallucination issues of decoder-only models—we introduce an auxiliary emotion classification objective. As illustrated in \cref{figure::avlm_emo_gen}, we extract hidden states corresponding to (1) visual query representations and (2) style/pitch tokens from the SpiritLM-Expressive Tokenizer (details in \cref{sec::exp_emo_gen}). These states are passed through two feedforward networks, concatenated, pooled, and classified using a cross-entropy loss. A stop-gradient is applied to prevent interference with AVLM training. During training, the true emotion label is used while at inference, the emotion label is predicted by the classifier.


\section{Experiments}
We begin with AVLM pre-training experiments (\cref{sec::exp_avlm_pretrain}), comparing different visual encoders and modality fusion strategies. Next, we fine-tune the AVLM on AVSR tasks (\cref{sec::exp_avsr}) to validate the effectiveness of our pre-training. Finally, we fine-tune AVLM for emotion recognition and expressive speech generation (\cref{sec::exp_emo_gen}), achieving superior performance over SpeechLM by utilizing visual cues.
\subsection{AVLM Pre-training}\label{sec::exp_avlm_pretrain}
\paragraph{Dataset and Preprocessing}
We use the LRS3 dataset \cite{afouras2018lrs3tedlargescaledatasetvisual}, which consists of 433 hours of transcribed audio-visual data, including a 30-hour clean subset. For pre-training, we utilize only the videos from the full 433-hour dataset. 


We adopt the data pre-processing pipeline from MuAViC \cite{anwar2023muavicmultilingualaudiovisualcorpus}, which segments videos into aligned audio-visual chunks of up to 15 seconds based on transcriptions. Speech tokens are extracted using SpiritLM’s tokenizer, which produces interleaved tokens of three types: semantic tokens (24.99 fps), style tokens (1 fps), and pitch tokens (12.5 fps). The semantic tokens are discretized HuBERT representations \cite{hsu2021hubert}, the pitch tokens are derived from a VQ-VAE model trained on the F0 of speech \cite{polyak2021speech}, and the style tokens are based on \textit{speechprop} features \cite{sonar_expressive}. For further details, we refer readers to \citet{nguyen2024spiritlm}.

\paragraph{Model Implementation Details}\quad We compare three visual encoders: (1) Open-MAGVIT2 \cite{magvit2}, (2) VGG-Face2 \cite{cao2018vggface2datasetrecognisingfaces}, and (3) SMIRK \cite{smirk}. Open-MAGVIT2 is a capable but expensive encoder as it is trained for image reconstruction. VGG-Face2 is more lightweight and focuses on facial regions through tasks like face recognition. SMIRK is a lightweight neural feature learned through 3D reconstruction based on FLAME \cite{flame}. We extract visual features using different encoders, adapt them with lightweight modules, and integrate them into the \textsc{Q-Former} for AVLM pre-training (visual encoder comparison details in \cref{app::visual_encoder_comp}). 

For all three fusion modules, we use the same base model, SpiritLM, and apply LoRA fine-tuning \cite{hu2021lora} to the query, key, value, and output projection matrices, with $r=16$, $\alpha=32$, and a dropout rate of 0.05. For \textsc{Q-Former Infill},  we replace 50\% of the speech tokens with visual query representations. For \textsc{Q-Former Prefix}, we drop attention over a certain percentage of speech tokens (from 0\% to 70\% as shown in \cref{table::noisy_ppl}) to see if such masking helps AVLM become more robust. The details and hyperparameters of our three fusion architectures are provided in \cref{app::avlm_implementation}.

\begin{table}[t]
  \centering
  \small
  \begin{tabular}{llc}
  \toprule
  \textbf{Fusion Module} & \textbf{Visual Feature} & \textbf{PPL} $\downarrow$ \\
  \midrule
  \multicolumn{3}{l}{\textit{Baseline}} \\
  \quad \textsc{Speech-only}            & N/A         & 5.6 \\
  \midrule
  \multicolumn{3}{l}{\textit{Fusion: Direct Concatenation}} \\
  \quad \textsc{Direct Concat} & SMIRK       & 132.0 \\
  \midrule
  \multicolumn{3}{l}{\textit{Fusion: Qformer Variants}} \\
  \quad \textsc{Q-Former Infill} & SMIRK       & 5.8 \\
  \quad \textsc{Q-Former Prefix} & SMIRK       & \textbf{5.5} \\
  \quad \textsc{Q-Former Prefix} & VGG-Face2    & 8.2 \\
  \quad \textsc{Q-Former Prefix} & MAGVIT2     & 8.6 \\
  \bottomrule
  \end{tabular}
  \caption{Perplexity (PPL) Comparison across Fusion Modules and Visual Features.}
  \label{table::fusion_visual_comparison}
  \vspace{-1em}
\end{table}

\noindent\textbf{Results}\quad We evaluate pre-training performance on the LRS3 test set using perplexity (PPL). First, fixing the fusion method to \textsc{Q-Former Prefix}, we ablate visual encoders. As shown in \cref{table::fusion_visual_comparison}, SMIRK achieves the lowest PPL, likely due to its disentangled expressive and jaw features being easier for the model to pick up. \textbf{We therefore adopt SMIRK as our default visual encoder for future fine-tuning experiments.}

Next, using SMIRK, we compare fusion strategies. \textsc{Q-Former Prefix} yields the lowest PPL, outperforming both \textsc{Speech-only} and other fusion methods. In contrast, \textsc{Direct Concat} performs worst,  suggesting that the direct concatenation and projection of audio-visual features complicate the model's adaptation to the representations.

Among Q-Former variants, \textsc{Q-Former Prefix} consistently performs best. We further evaluate robustness by applying attention masking at inference (10–70\%) and training additional \textsc{Q-Former Prefix} variants with various attention masking ratios (\cref{table::noisy_ppl}). A 30\% speech token dropout from attention computation during training yields the best overall results. Thus, we use the \textbf{\textsc{Q-Former Prefix} model trained with 30\% masking} as pre-trained base model for all fine-tuning tasks, including AVSR and expressive generation.

\begin{table}[t]
  \centering
  \small
  \resizebox{0.98\linewidth}{!}{
  \begin{tabular}{l cccccc}
    \toprule
    \textbf{Fusion Mode} & \textbf{Train} & \multicolumn{5}{c}{\textbf{PPL ($\downarrow$) @ Attention Mask Ratio}} \\
    \cmidrule(lr){3-7} & \textbf{Mask}                  & \textbf{0\%}  & \textbf{10\%} & \textbf{30\%} & \textbf{50\%} & \textbf{70\%} \\
    \midrule
    \textsc{Speech-only}                & --   & 5.56 & 6.02 & 7.69 & 10.27 & 13.62 \\
    \midrule
    \textsc{Infill}            & 50\% & 5.84 & 6.03 & 6.60 & 7.80 & 9.10 \\
    \midrule
    \textsc{Prefix}            & 0\%   &\textbf{5.5}& 5.87 & 7.46 & 9.81 & 12.0 \\
                              & 30\%  &5.65 & \textbf{5.81} & \textbf{6.31} & \textbf{7.50} & 8.88 \\
                                & 50\% &5.68 & 5.87 & 6.37 & 7.56 & 8.81 \\
                                & 70\%  &5.88 & 6.03 & 6.50 & 7.69 & \textbf{8.80} \\
    \bottomrule
  \end{tabular}}
  \caption{Perplexity under varying speech token mask ratios to compare the robustness of fusion modules. Each row is trained with a different speech token mask ratio.}
  \label{table::noisy_ppl}
  \vspace{-1em}
\end{table}

\begin{table*}[t]
  \centering
  \resizebox{0.95\textwidth}{!}{%
  \begin{tabular}{lcccccccccc}
    \toprule
    \multirow{2}{*}{\textbf{Model}} & \multirow{2}{*}{\textbf{Hours}} & \multirow{2}{*}{\textbf{Clean}} & \multicolumn{4}{c}{\textbf{WER (\%) @ SNR Noise Injection}} & \multicolumn{4}{c}{\textbf{WER (\%) @  Attention Mask Ratio}} \\
    \cmidrule(lr){4-7} \cmidrule(lr){8-11}
    & & & 10 dB & 5 dB & 2 dB & 0 dB & 10\% & 30\% & 50\% & 70\% \\
    \midrule
    \textsc{Speech-only} & 30 & 4.43 & 10.80 & 18.51 & 27.92 & 36.41 & 34.97 & 46.04 & 67.76 & 82.90 \\
    AVLM & 30 & 4.30 & 9.06 & 16.97 & 25.93 & 32.59 & 6.53 & 21.54 & 52.14 & 79.86 \\
    AVLM & 433 & \textbf{3.50} & \textbf{8.58} & \textbf{15.18} & \textbf{23.90} & \textbf{31.50} & \textbf{5.44} & \textbf{19.10} & \textbf{51.70} & \textbf{78.90} \\
    \bottomrule
  \end{tabular}
  }
  \vspace{-0.5em}
  \caption{WER (\%) of different models under varying SNR noise levels and speech token mask ratios.}
  \label{table::avsr}
  \vspace{-0.5em}
\end{table*}

\subsection{AVSR Fine-tuning}\label{sec::exp_avsr}

To further evaluate our pre-trained AVLM, we fine-tune it on the Audio-Visual Speech Recognition (AVSR) task using the 30-hour clean subset of LRS3 and report Word Error Rate (WER). We compare against two baselines: (1) a \textsc{Speech-only} SpiritLM model directly fine-tuned on the same 30-hour subset (using speech prefix and text continuation), and (2) an AVLM model with the same architecture as the pre-trained one but fine-tuned from scratch. 

AVSR serves as an auxiliary benchmark to assess the quality of visual integration beyond perplexity (used during pre-training). Due to space constraints, we omit modeling details, which largely mirror those in expressive generation (\cref{sec::emo_gen}) by using text transcriptions rather than response speech as targets. Full implementation details are in \cref{app::avsr_implementation}.

\begin{table}[t]
  \centering
  \small
  \setlength{\tabcolsep}{5pt}
  \renewcommand{\arraystretch}{1.2}
  \resizebox{0.99\linewidth}{!}{
  \begin{tabular}{lccc}
    \toprule
    \textbf{Model} & \textbf{Modality} & \textbf{Hours} & \textbf{WER (\%)} \\
    \midrule
    \multicolumn{4}{l}{\textit{Ours}} \\
    \quad \textsc{Speech-only (SpiritLM)} & S,T & 30 & 4.43 \\
    \quad \textsc{AVLM (Q-Former Prefix)} & S,T & 433 & 3.50 \\
    \midrule
    \multicolumn{4}{l}{\textit{Prior Work (Lip-only)}} \\
    \quad AV-HuBERT (Large) & T & 433 & 4.20 \\
    \quad MMS-LLaMa & T & 433 & 0.92 \\
    \quad LLaMa-AVSR & T & 433 & 0.95 \\
    \bottomrule
  \end{tabular}}
  \vspace{-0.5em}
  \caption{WER (\%) on clean speech. “Modality” indicates support for speech (S) or text (T) generation.}
  \label{table::avsr_comparison_toplines}
  \vspace{-1em}
\end{table}


We evaluate all models under both clean and noisy conditions. We introduce noise using two methods: (1) SNR noise injection and (2) attention masking. For SNR noise injection, we follow previous studies \cite{yeo2025mmsllamaefficientllmbasedaudiovisual} by adding white noise to create test audio with varying Signal-to-Noise Ratios (SNR). For attention masking, we randomly mask a certain percentage of speech tokens during attention computation.

\paragraph{Results} As demonstrated in \cref{table::avsr}, our pre-trained AVLM achieves the lowest Word Error Rate (WER) in both clean and noisy environments. This indicates that the visual integration via the \textsc{Q-Former Prefix} is effective, and pre-training on a larger dataset is advantageous. 

We also compare our results with previous top-performing models in \cref{table::avsr_comparison_toplines}. Although models like \rm{MMS-LLaMa} \cite{yeo2025mmsllamaefficientllmbasedaudiovisual} and \rm{LLaMa-AVSR} \cite{cappellazzo2025largelanguagemodelsstrong} report lower WER, the differences primarily stem from our choice of base model and visual features, rather than the modeling approach itself. 

Both \rm{MMS-LLaMa} and \rm{LLaMa-AVSR} utilize audio encoders like Whisper and focus on visual features from the lip region, limiting their models to the AVSR task. In contrast, our base model, SpiritLM, supports expressive speech generation by encoding speech into units, which inevitably loses information from speech tokenization and result in higher WER. Moreover, our visual features encompass the full face, not just the lip region, to facilitate expressive generation but at the cost of less effective semantic alignment for tasks like AVSR.

\begin{figure*}[t]
  \centering
  \includegraphics[width=0.95\linewidth]{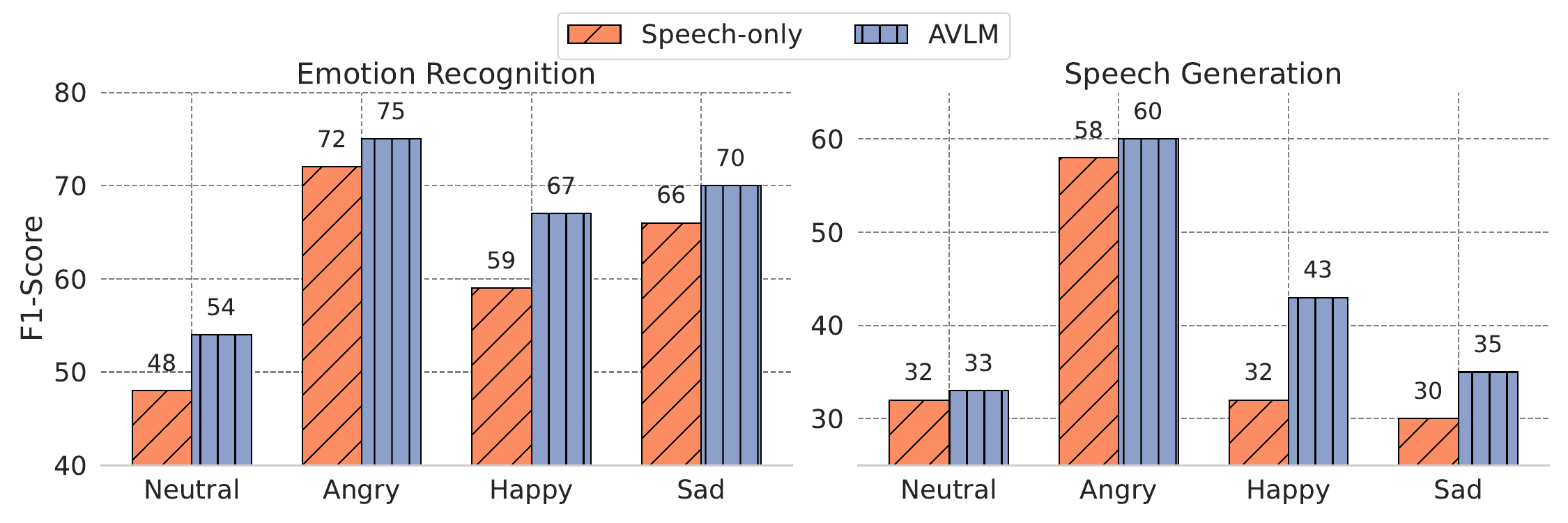}
  \vspace{-1em}
  \caption{F1-score of each emotion class for AVLM and \textsc{Speech-only} model.}
  \label{figure::emotion_f1_scores}
  \vspace{-1em}
\end{figure*}

\subsection{Emotion Recognition and Expressive Speech Generation}\label{sec::exp_emo_gen}
\paragraph{Dataset} 


Following the approach in \cref{sec::emo_gen}, we fine-tune our pre-trained AVLM using multimodal prompts consisting of visual input, input and response audio, and an emotion label. The dataset is built from IEMOCAP \cite{Busso2008IEMOCAPIE}, which provides expressive video dialogues and emotion labels. However, many original responses are very short (e.g., acknowledgments like “yeah”). To address this, we use GPT-4 \cite{openai2024gpt4}\footnote{We use the “gpt-4o-2024-11-20” snapshot} to rewrite conversations into longer, more detailed responses. We then generate corresponding audio using Step-Audio-TTS-3B \cite{stepaudio}, an expressive TTS model supporting voice cloning and style control. Additional dataset construction details are provided in \cref{app::dataset_details}.

\paragraph{Model Implementation}

We initialize our model from the pre-trained AVLM (\cref{sec::exp_avlm_pretrain}) using the \textsc{Q-Former Prefix} architecture with the SMIRK visual encoder, and fine-tune it with LoRA using the same hyperparameters as pre-training. The emotion classifier is trained separately on hidden states from visual queries and style/pitch tokens posii (\cref{sec::emo_gen}). For the speech-only baseline, we fine-tune a pre-trained SpiritLM model with LoRA, excluding visual inputs. Its emotion classifier uses only the style and pitch token hidden states.

\paragraph{Evaluation} 
For emotion recognition, we evaluate performance by comparing the predicted and ground truth emotion labels, computing both accuracy and F1-score for four emotion categories: Happy, Sad, Angry, and Neutral. For expressive speech generation, we use fine-tuned model to generate speech tokens and synthesize speech through SpiritLM's Tokenizer. Subsequently, we use a third-party model, Qwen2-Audio \cite{chu2024qwen2audiotechnicalreport}, which excels at audio understanding and emotion recognition, \textit{to predict emotion labels from the generated speech} (see \cref{app::avlm_implementation} for details). We then compute accuracy and F1-score for the same four-way classification. For reference, we have also uploaded audio of some generated responses in our ARR supplementary material.

\begin{table}[t]
  \centering
  \small
  \resizebox{0.98\linewidth}{!}{
  \begin{tabular}{lccc}
  \toprule
  \textbf{Model} & \textbf{UA (\%)} & \textbf{WA (\%)} & \textbf{F1 (\%)} \\
  \midrule
  \textit{Emotion Recognition} \\
  \quad \textsc{Speech-only} & 62.8 & 61.3 & 61.3 \\
  \quad \textsc{AVLM} & \textbf{67.2} & \textbf{67.1} & \textbf{66.2} \\
  \midrule
  \textit{Speech Generation} \\
  \quad \textsc{Speech-only} & 43.79 & 38.85 & 38.39 \\
  \quad \textsc{AVLM} & \textbf{46.03} & \textbf{42.99} & \textbf{42.49} \\
  \bottomrule
  \end{tabular}
  }
  \vspace{-0.5em}
  \caption{Performance comparison of AVLM and Speech-only model in terms of unweighted accuracy (UA), weighted accuracy (WA), and macro F1-score (F1).}
  \label{table::avlm_vs_speech_emo_recognition}
  \vspace{-1em}
\end{table}

\paragraph{Results}
As shown in \cref{table::avlm_vs_speech_emo_recognition}, our AVLM model outperforms the \textsc{Speech-only} model on both emotion recognition and speech generation tasks. Note that, both models achieve better performance on emotion recognition, which is expected since this task directly optimizes for emotion label prediction. In contrast, speech generation is more challenging: the decoder-only model may hallucinate, producing speech responses that fail to convey the intended emotion. Nevertheless, across both tasks, our AVLM model consistently surpasses the \textsc{Speech-only} baseline, highlighting the benefit of visual guidance. In \cref{figure::emotion_f1_scores}, we further break down the F1-scores by emotion category, observing a similar trend where AVLM always outperforms its speech-only counterpart. Among the emotions, Neutral and Sad are generally harder to predict than Happy and Angry, likely because they convey weaker emotional cues. On the contrary, Happy and Angry are associated with stronger facial expressions and higher vocal arousal, making it easier for the model to predict or generate speech of similar properties.

\paragraph{Emotion Controllability Analysis} We investigated whether the emotion of generated speech could be controlled by modifying the predicted emotion label in the prompt. For example, if the original predicted emotion is “angry”, we manually change it to “happy” and then evaluate the emotion of the newly generated speech. We measure the number of instances where the generated emotion successfully shifts from the original to the altered label and visualize the results in a heatmap. Ideally, we expect our AVLM to demonstrate strong controllability, generating speech that appropriately reflects the emotion specified in the prompt.

\begin{figure}[t]
  \centering
  \includegraphics[width=1.05\linewidth, trim={1cm 0 0 0}, clip]{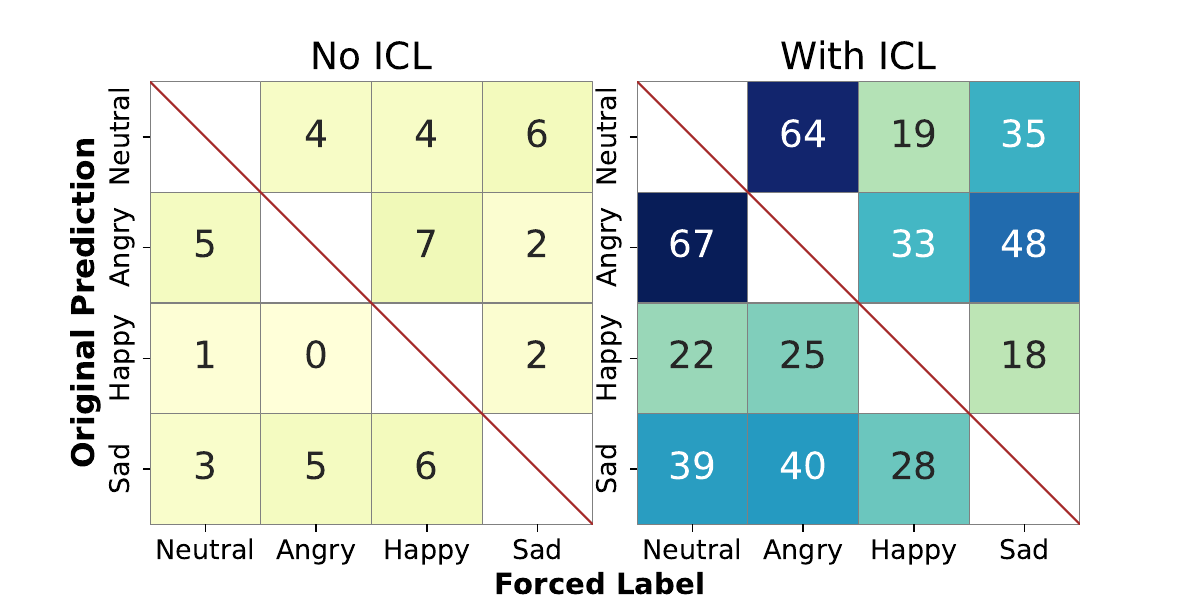}
  \caption{Heatmaps of emotion controllability analysis for AVLM. The number indicates the amount of generated speech that changes from its original emotion to the forced emotion label. ICL: In-context Learning.}
  \vspace{-1em}
  \label{figure::emotion_control_heatmaps}
\end{figure}

As shown in \cref{figure::emotion_control_heatmaps}, simply changing the emotion label in the prompt—without using in-context learning—rarely alters the emotional tone of the generated speech. This suggests that the model primarily relies on input speech and visual cues to guide emotion, rather than the emotion label itself. This behavior is expected, given the limited training data and the fact that input and response audio in our dataset are designed to share the same emotion.

To enhance controllability, we introduce in-context learning by including one demonstration per emotion in the prompt (see example in \cref{app::avlm_implementation}). With this addition, AVLM becomes more responsive to changes in the emotion label, as shown in the right panel of \cref{figure::emotion_control_heatmaps}.

These findings indicate there is still large room for improving the emotion controllability, as the model after fine-tuning is not able to easily alter the speech style by conditioning on different emotion labels. We believe that enhancing data quality by creating larger and more diverse expressive audio-visual dialogue is crucial to addressing this issue, and we plan to explore this in future work.

\section{Conclusion}
We present a framework for Audio-Visual Language Modeling that effectively incorporates visual cues into SpeechLM. Following empirical exploration of visual encoders and modeling architectures, we employ a prefix-based Query-Former with SMIRK visual features for AVLM. Our system, through leveraging visual modality, outperforms speech-only baselines in both Audio-Visual Speech Recognition and Expressive Speech Generation.

\section*{Acknowledgement}
We express our profound appreciation to anonymous reviewers for their helpful suggestions. We also thank Yue Xu, Tianjian Li,Jingyu Zhang for their valuable suggestions to enhance the presentation of this work.

\section*{Limitations and Broader Impact}
One limitation of our work is the limited availability of expressive audio-visual dialogue data for fine-tuning the pre-trained AVLM. Although we synthesize part of our dataset, the construction pipeline is based on IEMOCAP, which contains only a few hours of recordings. Collecting more diverse and expressive dialogue data could further enhance the controllability and quality of the generated speech.

Another limitation is that our evaluation primarily focuses on the emotional expressiveness of the generated speech, reflecting our goal of modeling emotion-aware speech generation from audio-visual inputs. While we demonstrate that visual cues can significantly improve emotional alignment, other aspects of speech quality—such as helpfulness, factual accuracy, and coherence—are not evaluated in this study. Finally, the decoding process in our \textsc{Q-Former Prefix} architecture requires processing the visual stream before the speech stream, which can introduce latency and leave room for further efficiency improvements.

\textbf{Ethical Consideration}: While our work advances the generation of emotionally expressive and natural-sounding speech, it also raises potential risks of misuse. Specifically, models capable of synthesizing human-like emotional speech may be exploited for deceptive purposes, such as impersonation, social engineering scams, or the spread of misinformation. To mitigate such concerns, we encourage future work to incorporate safeguards such as synthetic speech watermarking, controlled access to fine-tuned models, and explicit labeling of machine-generated content. Responsible deployment and continuous risk assessment are essential as these technologies mature.

Our experiments are conducted on publicly available datasets under appropriate licenses. The LRS3 dataset is released under the Creative Commons Attribution 4.0 International License and is available for research purposes. The IEMOCAP dataset is also released for academic research use under its license agreement. We use both datasets in accordance with their respective terms to ensure compliance with data usage policies. The synthesized artifacts for fine-tuning are not currently distributed to avoid potential misuse and to respect the licensing constraints from IEMOCAP \citep{Busso2008IEMOCAPIE}. We are planning to release the dataset if permission is granted from the IEMOCAP data provider. We believe responsible handling of such synthetic content is critical to ensuring ethical standards in expressive speech research.

We also acknowledge the use of AI assistants (e.g., GitHub Copilot, ChatGPT) to support the development of our research. These tools were used to assist with code implementation, debugging, documentation, and drafting or refining written content. All substantive research contributions, model design decisions, and experimental results were generated and verified by the authors. We ensured that the use of AI tools complied with academic integrity standards and that any generated content was critically reviewed and edited to maintain originality and correctness.

\bibliography{custom}

\newpage
\onecolumn
\appendix
\section*{\LARGE{Supplementary Material}}\label{sec:appendix}

\begin{table}[ht]
    \centering
    \small
    \begin{tabular}{cl}
    \textbf{Appendix Sections}    & \textbf{Contents}  \\ \toprule
    \autoref{app::dataset_details}     &  \begin{tabular}[c]{@{}l@{}} Expressive Conversation Dataset Construction from IEMOCAP\end{tabular}\\ \midrule
    \autoref{app::iemocap_emo_recognition} &  \begin{tabular}[c]{@{}l@{}} IEMOCAP Emotion Recognition Experiment Setup \end{tabular} \\ \midrule
    \autoref{app::visual_encoder_comp} &  \begin{tabular}[c]{@{}l@{}} Comparison of Visual Encoders for AVLM Pre-training \end{tabular} \\ \midrule
    \autoref{app::avlm_implementation} &  \begin{tabular}[c]{@{}l@{}} Audio-Visual Language Modeling Implementation Details \end{tabular} \\ \midrule
    \autoref{app::avsr_implementation} &  \begin{tabular}[c]{@{}l@{}} Audio-Visual Speech Recognition Modeling Details \end{tabular} \\
    \bottomrule
\end{tabular}    
\end{table}

\section{Expressive Audio-Visual Dataset Construction}\label{app::dataset_details}
To showcase the benefit of our visual integration, we synthesize data from IEMOCAP \cite{Busso2008IEMOCAPIE} to create an expressive data collection that contains video input, response audio, and emotion label. Note that although the original IEMOCAP dataset already contains data format that satisfies this requirement, we find the conversation are not good enough to fine-tune our AVLM because a lot of the dialogue turns are too short to be useful. Some example conversation from original dataset are shown below in \cref{tab:original-dialogues}.

\begin{table}[ht]
  \centering
  \small
  \begin{tabular}{c p{12cm} p{2cm}}
    \toprule
    \textbf{Example} & \textbf{Speaker A} & \textbf{Speaker B} \\
    \midrule
    1 & Do you have your forms? & Yeah. \\
    2 & Who told you to get in this line? & You did. \\
    3 & I'm a big fan of the Wheel of Fortune quarters. But it just costs so much. I don't know. & Oh. \\
    4 & Well so now we're married. We have cat children. & Awesome. \\
    \bottomrule
  \end{tabular}
  \caption{Original short conversations motivating synthetic augmentation}
  \label{tab:original-dialogues}
\end{table}

In order to fine-tune our AVLM for better response generation, we perform conversatoin re-writing to generate more detailed response using GPT-4 \cite{openai2024gpt4}. The prompt we use is following:

\begin{tcolorbox}[
  colback=brown!5!white,
  colframe=brown!75!black,
  title=Conversation Rewriting Prompt,
  fonttitle=\footnotesize\bfseries,
  sharp corners,
  boxrule=0.8pt,
  left=2mm, right=2mm, top=1mm, bottom=1mm
]
\footnotesize
\ttfamily
\noindent
{\color{black}<Instruction>:} {\color{grayblack}Your task is to continue the following conversation naturally, taking into account both the conversation history and the provided emotional tone. You will generate a response from the perspective of a} {\color{blue}\textit{[speaker\_gender]}} {\color{grayblack}speaker, ensuring that your tone reflects a} {\color{purple}\textit{[emotion]}} {\color{grayblack}emotion. An example response is also provided for reference; if it is very short, please expand it to one or two sentences.}\\[2pt]

{\color{black}<Conversation History>:} {\color{blue}\textit{[Previous Conversation]}}

{\color{black}<Current Speaker's Emotion>:} {\color{purple}\textit{[Emotion Label]}}

{\color{black}<Example Response>:} {\color{redbrown}\textit{[Original Response Text]}}\\

{\color{grayblack}Now, generate a natural and concise response that continues the dialogue using all of the above information. 
Keep your answer to approximately two sentences. Output the response only.}
\end{tcolorbox}

The rewritten conversation will have better length control as we prompt the model to keep the response to approximately two sentences. Since the response also takes into account all conversation history from previous turns and the current speaker gender and emotion, it will greatly improve the quality of the dialogue for training our AVLM. Then we need to synthesize audio from the re-written conversations and we choose Step-Audio-TTS-3B \cite{stepaudio} to because it supports voice cloning and expressive speech synthesis with natural language style control. We feed the autoregressive TTS model with the following prompt: \texttt{[Emotion] [Text Response]} along with a speaker voice clip (from the current conversation) for the model to clone.

After the synthesis process above, we get 4859 conversation pairs and we split it by 80:10:10 for train:dev:test. During our construction, we use five emotion labels: Happy, Sad, Angry, Neutral, and Frustrated, as these are the most common emotions from the original dataset, and we filter out all conversation with other emotion labels. However, we also realize that the emotion Angry and Frustrated are very similar so we merge them into one class to only have Angry emotion. Lastly, we process the visual stream by first cropping the face region with YOLO \cite{redmon2016lookonceunifiedrealtime}\footnote{~We use publicly available ckpt: yolov8l-face-lindevs.pt} and then extract its SMIRK feature. We process the audio stream by SpiritLM's expressive tokenizer to produce semantic, style, and pitch tokens. An example of crafted data is shown in \cref{table::all-processed-example}.

\begin{table}[ht]
  \centering\small
  \begin{tabularx}{\linewidth}{@{}lY@{}}
    \toprule
    \textbf{Field} & \textbf{Content} \\
    \midrule

    \multicolumn{2}{@{}l}{\textbf{Example A}} \\ 
    \addlinespace[0.5ex]
    Emotion        & \color{asuffix!80!black}{Sad} \\
    Input Visual   & SMIRK features extracted at 25 fps \\
    Input Text     & No, I know but—I know I don't make you happy. \\
    Response Text  & It's just that sometimes it feels like we're stuck in this place, and I don't know how to find my way back to feeling good again. I wish things were different, but I don't want to push you away either. \\
    Input Units    & {\scriptsize[\styleToken{St48}]\allowbreak[\pitchToken{Pi40}]\allowbreak[Hu109]\allowbreak[Hu118]\allowbreak[Hu80]\allowbreak[\pitchToken{Pi44}]\allowbreak[Hu42]\allowbreak[Hu112]\allowbreak[Hu342]\allowbreak[Hu490]\allowbreak[Hu356]\allowbreak[\styleToken{St48}]\allowbreak[Hu136]\allowbreak[\pitchToken{Pi56}]\allowbreak[Hu1]\allowbreak[Hu94]\allowbreak[\pitchToken{Pi44}]\allowbreak[Hu114]...} \\
    Response Units & {\scriptsize[\styleToken{St28}]\allowbreak[\pitchToken{Pi9}]\allowbreak[Hu140]\allowbreak[\pitchToken{Pi37}]\allowbreak[Hu482]\allowbreak[Hu465]\allowbreak[Hu178]\allowbreak[\pitchToken{Pi21}]\allowbreak[Hu7]\allowbreak[Hu132]\allowbreak[\pitchToken{Pi49}]\allowbreak[Hu397]\allowbreak[\pitchToken{Pi8}]\allowbreak[Hu145]\allowbreak[Hu248]\allowbreak[Hu493]\allowbreak[Hu383]\allowbreak[Hu30]...} \\

    \addlinespace[1ex]
    \midrule
    \multicolumn{2}{@{}l}{\textbf{Example B}} \\ 
    \addlinespace[0.5ex]
    Emotion        & \color{asuffix!80!black}{Angry} \\
    Input Visual   & SMIRK features extracted at 25 fps \\
    Input Text     & Do you want to – do you want to leave us behind? I mean, I don't understand why you have to do this. \\
    Response Text  & I wish it were different, but this is something I committed to long before any of this happened. It's not about wanting to leave you behind; it's about fulfilling my responsibility. \\
    Input Units    & {\scriptsize[\styleToken{St71}]\allowbreak[\pitchToken{Pi21}]\allowbreak[Hu267]\allowbreak[\pitchToken{Pi52}]\allowbreak[Hu7]\allowbreak[\pitchToken{Pi0}]\allowbreak[Hu469]\allowbreak[Hu118]\allowbreak[\pitchToken{Pi47}]\allowbreak[Hu410]\allowbreak[Hu333]\allowbreak[Hu156]\allowbreak[\pitchToken{Pi21}]\allowbreak[Hu139]\allowbreak[\pitchToken{Pi9}]\allowbreak[Hu359]\allowbreak[\pitchToken{Pi5}]\allowbreak[Hu343]...} \\
    Response Units & {\scriptsize[\styleToken{St71}]\allowbreak[\pitchToken{Pi41}]\allowbreak[Hu255]\allowbreak[Hu437]\allowbreak[Hu360]\allowbreak[Hu35]\allowbreak[\pitchToken{Pi46}]\allowbreak[Hu169]\allowbreak[\pitchToken{Pi25}]\allowbreak[Hu70]\allowbreak[Hu431]\allowbreak[\pitchToken{Pi52}]\allowbreak[Hu389]\allowbreak[Hu249]\allowbreak[\pitchToken{Pi54}]\allowbreak[Hu95]...} \\
    \addlinespace[1ex]
    \midrule
    \multicolumn{2}{@{}l}{\textbf{Example C}} \\ 
    \addlinespace[0.5ex]
    Emotion        & \color{asuffix!80!black}{Happy} \\
    Input Visual   & SMIRK features extracted at 25 fps \\
    Input Text     & Great. That makes us all happy here at D.S.L. Extreme.  Is there anything else I can help you with? \\
    Response Text  & No, I think that's everything for now! You've really made my day brighter. I'm excited to stick with D.S.L. Extreme! \\
    Input Units    & {\scriptsize[\styleToken{St22}]\allowbreak[\pitchToken{Pi59}]\allowbreak[Hu7]\allowbreak[\pitchToken{Pi20}]\allowbreak[Hu171]\allowbreak[\pitchToken{Pi59}]\allowbreak[Hu118]\allowbreak[Hu161]\allowbreak[\pitchToken{Pi48}]\allowbreak[Hu71]\allowbreak[Hu142]\allowbreak[\pitchToken{Pi46}]\allowbreak[Hu251]\allowbreak[Hu175]\allowbreak[\pitchToken{Pi41}]\allowbreak[Hu7]\allowbreak[Hu314]\allowbreak[\pitchToken{Pi40}]...} \\
    Response Units & {\scriptsize[\styleToken{St69}]\allowbreak[\pitchToken{Pi2}]\allowbreak[Hu278]\allowbreak[Hu80]\allowbreak[\pitchToken{Pi54}]\allowbreak[Hu112]\allowbreak[Hu81]\allowbreak[\pitchToken{Pi52}]\allowbreak[Hu159]\allowbreak[\pitchToken{Pi11}]\allowbreak[\pitchToken{Pi30}]\allowbreak[Hu289]\allowbreak[Hu126]\allowbreak[\pitchToken{Pi2}]\allowbreak[Hu362]\allowbreak[\pitchToken{Pi30}]\allowbreak[Hu56]\allowbreak[\pitchToken{Pi2}]\allowbreak[Hu7]...} \\

    \addlinespace[1ex]
    \midrule
    \multicolumn{2}{@{}l}{\textbf{Example D}} \\ 
    \addlinespace[0.5ex]
    Emotion        & \color{asuffix!80!black}{Neutral} \\
    Input Visual   & SMIRK features extracted at 25 fps \\
    Input Text     & She's actually from England, I think.  Well she- she spoke English but she had an accent. I didn't even ask. \\
    Response Text  & I guess that works out for both of you then. Have you thought about what it would be like to actually go through with the marriage? \\
    Input Units    & {\scriptsize[\styleToken{St28}]\allowbreak[\pitchToken{Pi57}]\allowbreak[Hu475]\allowbreak[\pitchToken{Pi20}]\allowbreak[\pitchToken{Pi35}]\allowbreak[\pitchToken{Pi47}]\allowbreak[Hu28]\allowbreak[\pitchToken{Pi54}]\allowbreak[\pitchToken{Pi52}]\allowbreak[\pitchToken{Pi47}]\allowbreak[Hu49]\allowbreak[\pitchToken{Pi35}]\allowbreak[Hu391]\allowbreak[Hu245]\allowbreak[\pitchToken{Pi13}]\allowbreak[Hu227]\allowbreak[Hu355]...} \\
    Response Units & {\scriptsize[\styleToken{St71}][\pitchToken{Pi20}][Hu278][Hu35][\pitchToken{Pi35}][Hu130][\pitchToken{Pi30}][Hu169][\pitchToken{Pi56}][Hu459][Hu131][\pitchToken{Pi47}][Hu197][Hu128][\pitchToken{Pi5}][Hu424][\pitchToken{Pi11}][Hu368]...} \\

    \bottomrule
  \end{tabularx}
  \caption{Examples of processed data entries of different emotions. The input text and response text are only shown for reference but not used during AVLM fine-tuning experiments.}
  \label{table::all-processed-example}
\end{table}

The synthesized artifacts for fine-tuning are not currently distributed to avoid potential misuse and to respect the licensing constraints from IEMOCAP \citep{Busso2008IEMOCAPIE}. We are planning to release the dataset if permission is granted from the IEMOCAP data provider. We believe responsible handling of such synthetic content is critical to ensuring ethical standards in expressive speech research.


\section{IEMOCAP Emotion Recognition Experiment Setup}\label{app::iemocap_emo_recognition}
In \cref{sec::iemocap_recognition}, we demonstrate the benefit of incorporating visual modality for emotion recognition. Below, we detail our experimental setup:

We follow the data split protocol from EmoBox \cite{Ma2024EmoBoxMM}, using the publicly available splits at \url{https://github.com/emo-box/EmoBox/tree/main/data/iemocap}. Visual frames are encoded using the Open-MAGVIT2 encoder, while speech signals are processed with \textit{Whisper-Large-V3}. A lightweight adaptation module is then applied to map both speech and visual features into 512-dimensional representations for each frame.

To adapt the visual features from Open-MAGVIT2, which produce tensors of shape $(Z=18, H=32, W=32)$, we stack three residual blocks, each consisting of a normalization layer followed by a 2D convolutional neural network (CNN). The resulting output is flattened and projected into a 512-dimensional feature vector. Speech features are adapted in a similar manner, but using 1D-CNNs instead of 2D-CNNs.

The adapted speech and visual features are then concatenated and fed into a Transformer encoder consisting of 6 layers with a hidden size of 512. We experimented with concatenating along both the temporal and feature dimensions but observed no significant difference in performance. Therefore, we adopt feature-dimension concatenation to keep the sequence length shorter. The resulting features are mean-pooled and passed through a two-layer feedforward network to predict the emotion label. We train the model until convergence, using a learning rate of $3 \times 10^{-5}$, a batch size of 16, and gradient accumulation over 4 steps.

To evaluate classifier performance, we follow prior work \cite{Ma2024EmoBoxMM, sensevoice} and report the unweighted average accuracy (UA), weighted average accuracy (WA), and macro F1 score, as shown in \cref{table::audio_visual_emo_classification}. The UA, WA, and F1 metrics are computed using the \textit{accuracy\_score} and \textit{balanced\_accuracy\_score} functions from the \textit{sklearn.metrics} package.

\section{Visual Encoder Comparison}\label{app::visual_encoder_comp}
We initially selected MAGVIT2 (using the public checkpoint from Open-MAGVIT2) for its rich visual representations. However, MAGVIT2 encodes each $256 \times 256$ frame into a high-dimensional feature of size $18 \times 32 \times 32 = 18{,}432$, with 18 channels and spatial dimensions downsampled by a factor of 8. This high dimensionality introduces significant computational overhead and poses learning challenges, particularly given the limited size of our dataset (fewer than 500 hours from LRS3). To mitigate these issues, we explored more lightweight visual encoders: VGG-Face2 and SMIRK.

VGG-Face2, trained for face recognition, encodes each frame into a 512-dimensional vector. SMIRK, trained for 3D face reconstruction using FLAME \cite{flame}, disentangles facial attributes into components such as shape, expression, jaw, eyelid, pose, and camera parameters. Among these, we use only the expressive (55-dimensional) and jaw (3-dimensional) features, which are most relevant for capturing emotion and lip movements.

Since the outputs of these visual encoders differ in shape, we design specialized adapters before feeding the features into the \textsc{Q-Former}. For MAGVIT2, we follow the setup in \cref{app::iemocap_emo_recognition}, stacking 2D convolutional layers with residual connections to reduce the spatial dimensions and project the encoding into a 512-dimensional feature.

For VGG-Face2, whose outputs are already 512-dimensional, we apply a simple two-layer feedforward network that first expands the feature to 1024 dimensions before projecting it back to 512 dimensions. For SMIRK, we linearly transform the expressive parameters into a 128-dimensional vector and the jaw parameters into a 32-dimensional vector. These are concatenated and passed through a two-layer feedforward network to produce a 256-dimensional visual representation.

Since SMIRK features can sometimes be distorted, we leverage its pose parameters to filter out video clips with excessive horizontal head rotation. Let the predicted pose vector be $\bm{r}_{\text{pose}} = (r_x, r_y, r_z)$, initially represented as a rotation vector. We convert it into a rotation matrix $\bm{R}$ using Rodrigues' rotation formula:
\begin{equation}
\label{eq::rodrigues_rotation}
\bm{R} = \mathbf{I} + \sin(\theta) \mathbf{K} + (1 - \cos(\theta)) \mathbf{K}^2
\end{equation}
where $\theta = |\bm{r}_{\text{pose}}|$ and $\mathbf{K}$ is the skew-symmetric matrix of the unit vector $\mathbf{u} = \frac{\bm{r}_{\text{pose}}}{\theta}$, with $\mathbf{I}$ denoting the identity matrix. The resulting matrix $\bm{R}$ is then converted to Euler angles $(\phi, \theta, \psi)$, corresponding to yaw, pitch, and roll. We use the yaw angle $$\phi = \text{arctan2}(-R_{31}, R_{33})$$ (where $R_{ij}$ are the elements of the rotation matrix $\bm{R}$) to quantify horizontal head rotation and filter out clips whose average yaw angle exceed 30\%, resulting in about 10\% drop in the training data.

As shown in \cref{table::fusion_visual_comparison}, we compare the three visual encoders using the \textsc{Q-Former Prefix} architecture and observe that SMIRK outperforms both MAGVIT2 and VGG-Face2. We attribute this superior performance to SMIRK’s ability to produce disentangled and emotionally relevant features, thereby facilitating more effective learning.

\section{Audio-Visual Language Modeling Implementation Details}\label{app::avlm_implementation}

\paragraph{Training}
We explored three modality fusion strategies. In the \textsc{Direct Concat} approach, the encoded visual features are first projected to 256 dimensions and then concatenated with the 4096-dimensional speech features from SpiritLM, resulting in a combined 4352-dimensional representation. This concatenated vector is subsequently projected back to 4096 dimensions through a two-layer feedforward network.

For the \textsc{Q-Former Prefix} and \textsc{Q-Former Infill} methods, we leverage query latents to retrieve relevant visual information using a 6-layer cross-attention module. Before computing cross-attention, we add positional embeddings to the query latents to encode temporal information. In \textsc{Q-Former Prefix}, we apply a compression ratio of 5 to downsample the visual input (\ie for a one-second video at 25 frames per second, we generate 5 query representations to serve as the prefix). In contrast, in \textsc{Q-Former Infill}, the query latents are sized to match the number of speech tokens; after retrieval from the visual stream, 50\% of the speech representations are replaced with the retrieved visual queries.

We adopt LoRA for training our AVLM, using hyperparameters $r=16$, $\alpha=32$, and a dropout rate of 0.05. The LoRA parameters are optimized with a learning rate of $3 \times 10^{-5}$ and 1000 warm-up steps. For training the visual adapter (described in \cref{app::visual_encoder_comp}), we use a larger learning rate of $1 \times 10^{-4}$. The model is trained with a batch size of 8, applying gradient accumulation every 2 steps.

For AVLM fine-tuning on speech generation, we adjust the batch size to 5 due to memory constraints. For the emotion classifier, we train it with learning rate $3e^{-4}$ and for the LoRA parameters, we use learning rate $5e^{-5}$ with no warmup steps. Throughout all experiments, we use $4\times$ A100 80G GPUs and apply Adam Optimizer with $\beta_1 = 0.99, \beta_2 = 0.999$.

\paragraph{Inference}
During inference, we prepare a multimodal prompt using the text instruction, visual input, and audio input as shown in the zero-shot prompt below:

\begin{tcolorbox}[
  colback=brown!5!white,
  colframe=brown!75!black,
  title=Zero-Shot Prompt,
  fonttitle=\footnotesize\bfseries,
  sharp corners,
  boxrule=0.8pt,
  left=2mm, right=2mm, top=1mm, bottom=1mm,
]
\footnotesize
\ttfamily
\noindent
<Text Instruction>: {\color{gray}Perceive the given visual and audio input. Determine the emotion of the speaker and continue the dialogue with the same emotion}.\\[2pt]
<Visual Input>: {\color{yellow!80!black} $\bm{z}_1, \bm{z}_2, \cdots, \bm{z}_N$}\\
<Input Audio>: {\color{blue!80!black} $\bm{s}_1, \bm s_2, \cdots, \bm s_T$}
\end{tcolorbox}

Then, we infer the emotion from our trained audio-visual emotion classifier module, and insert the emotion into the prompt. Lastly, the AVLM model generates speech continuation based on this multimodal input prompt.

For In-context Learning (ICL) analysis, we use a different prompt that contains an example of each emotion, as shown in the prompt below. As discussed in \cref{sec::exp_emo_gen}, with prompt demonstrations, the generated speech better aligns with the emotion label compared to zero-shot prompting, indicating that the model can learn from context to adapt its emotion during expressive generation. This also suggests room for further improvements, as zero-shot setting is not able to achieve good emotion-controllability itself.

\begin{tcolorbox}[
  colback=brown!5!white,
  colframe=brown!75!black,
  title=In-context Learning Prompt,
  fonttitle=\footnotesize\bfseries,
  sharp corners,
  boxrule=0.8pt,
  left=2mm, right=2mm, top=1mm, bottom=1mm,
]
\footnotesize
\ttfamily
\noindent
<Text Instruction>: {\color{gray}Perceive the given visual and audio input. Determine the emotion of the speaker and continue the dialogue with the same emotion. Below are some examples:}\\[2pt]\\
<Audio-Visual Conversation Input>: {\color{yellow!80!black}visual queries};{\color{blue!80!black}input speech units}\\
<Emotion>: {\color{asuffix!80!black}happy}\\
<Response>: {\color{green!50!black}response speech units}\\
\\
<Audio-Visual Conversation Input>: {\color{yellow!80!black}visual queries};{\color{blue!80!black}input speech units}\\
<Emotion>: {\color{asuffix!80!black}sad}\\
<Response>: {\color{green!50!black}response speech units}\\
\\
<Audio-Visual Conversation Input>: {\color{yellow!80!black}visual queries};{\color{blue!80!black}input speech units}\\
<Emotion>: {\color{asuffix!80!black}angry}\\
<Response>: {\color{green!50!black}response speech units}\\
\\
<Audio-Visual Conversation Input>: {\color{yellow!80!black}visual queries};{\color{blue!80!black}input speech units}\\
<Emotion>: {\color{asuffix!80!black}neutral}\\
<Response>: {\color{green!50!black}response speech units}\\
\\
{\color{gray}Now continue the following Audio-Visual conversation input with specified emotion}\\
<Visual Input>: {\color{yellow!80!black} $\bm{z}_1, \bm{z}_2, \cdots, \bm{z}_N$}\\
<Input Audio>: {\color{blue!80!black} $\bm{s}_1, \bm s_2, \cdots, \bm s_T$}\\
<Emotion>: {\color{asuffix!80!black} \{emotion label\}}\\
<Response>: 
\end{tcolorbox}

To control the structure of generated speech token sequences, we employ a custom decoding strategy using nucleus sampling with temperature, combined with a logits processor. Specifically, we set the generation temperature to 0.8 and use top-$p$ sampling with $p=0.95$, generating up to 300 new tokens with sampling enabled. Additionally, we apply a LogitProcessor to enforce structural constraints on the token types during generation. 

The processor masks out all non-speech tokens and selectively restricts transitions between different token types. Speech tokens are grouped into three disjoint ranges: style tokens, pitch tokens, and semantic tokens. At the first decoding step, only style tokens are allowed. Thereafter, we prohibit consecutive style or pitch tokens by dynamically masking their respective ranges based on the previously generated token type. No such constraint is applied to semantic tokens, allowing them to appear in sequence. This decoding logic helps the model maintain a well-formed token sequence that can be converted back to expressive speech waveforms through pre-trained speech decoders.

To detect the emotion from the generated speech, we perform evaluation with Qwen2-Audio. We prompt it with following text: \textit{What's the emotion of the audio? Only output the emotion label from the following list: Happy, Sad, Angry, Frustrated, Neutral} and provide the synthesized audio. The predicted labels are then used to compute accuracy and F1-score against the ground-truth emotion labels.

\section{AVSR Finetuning Details}\label{app::avsr_implementation}

\begin{figure*}[t]
  \centering
  \includegraphics[width=0.9\linewidth]{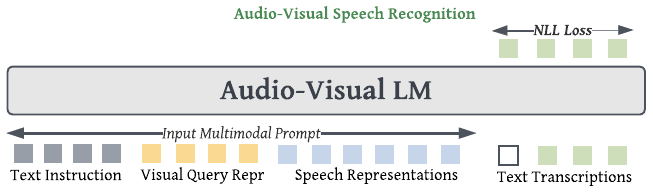}
  \caption{Audio-Visual Speech Recognition Fine-tuning based on pre-trained AVLM model.}
  \vspace{-1em}
  \label{figure::avsr_model}
\end{figure*}

In this section, we elaborate on the modeling and experimental setup used for the AVSR (Audio-Visual Speech Recognition) task, which was omitted from the main text due to space limitations. Our fine-tuning strategy mirrors the multimodal prompting approach described in \cref{sec::exp_emo_gen}, but with one key difference: instead of generating a response in speech, the model is trained to predict transcribed text tokens, as illustrated in \cref{figure::avsr_model}. The instruction-tuning prompt used for AVSR adapts accordingly to reflect this new objective. For the speech-only baseline, we directly fine-tune SpiritLM with the same multimodal prompt, except that visual representations are not provided.
\begin{tcolorbox}[
  colback=brown!5!white,
  colframe=brown!75!black,
  title=Instruction-Tuning Prompt,
  fonttitle=\footnotesize\bfseries,
  sharp corners,
  boxrule=0.8pt,
  left=2mm, right=2mm, top=1mm, bottom=1mm,
]
\footnotesize
\ttfamily
\noindent
<Text Instruction>: {\color{gray}Transcribe the following audio visual clip:}\\[2pt]
<Visual Input>: {\color{yellow!80!black} $\bm{z}_1, \bm{z}_2, \cdots, \bm{z}_N$}\\
<Audio Input>: {\color{blue!80!black} $\bm{s}_1, \bm s_2, \cdots, \bm s_T$}\\
<Transcription>: {\color{green!50!black}(some text tokens)}
\end{tcolorbox}



We fine-tune our AVLM on the LRS3 dataset using LoRA-based parameter-efficient updates. The training configuration aligns closely with that used for expressive speech generation. Specifically, for the baseline AVLM variant without pretraining, we apply a learning rate of $1 \times 10^{-4}$ to the fusion modules and $3 \times 10^{-5}$ to the LoRA parameters within SpeechLM. In contrast, for the pretrained AVLM, we freeze the fusion modules and update only the LoRA parameters. Fine-tuning is conducted with a batch size of 24, using gradient accumulation over 8 steps and 1,000 warmup steps.

To encourage the model to leverage visual input, we adopt a speech masking strategy during both training and evaluation. For a specified dropout ratio, we randomly sample a subset of positions from the input speech token sequence and prevent attention from attending to those positions. This forces the model to rely more heavily on visual cues when performing recognition (results shown in \cref{table::avsr}).

\end{document}